\documentclass{article}
\usepackage{spconf,amsmath,graphicx,hyperref}

\usepackage{silence}
\usepackage{amsmath,amssymb,amsfonts}
\usepackage{soul}
\usepackage[table]{xcolor}
\usepackage{multirow}
\usepackage{subcaption}
\usepackage{multirow,calc,bm,diagbox,makecell,hhline}
\usepackage[nopartial]{fixdif}

\usepackage{array}\makeatletter
\newcommand{\@Deccolumntype}[1]{\let\@temp=\\#1\let\\=\@temp}
\newcolumntype{C}[1]{>{\@Deccolumntype\centering}p{#1}}
\newcolumntype{R}[1]{>{\@Deccolumntype\raggedleft}p{#1}}
\newcolumntype{L}[1]{>{\@Deccolumntype\raggedright}p{#1}}
\newcolumntype{\|}{@{\hskip\tabcolsep\vrule width 1.0pt\hskip\tabcolsep}}
\makeatother
\def\Pr{\operatorname*{P}}% \def\Pr{\mathdif{\mathrm P}}
\def\leq{\leqslant}
\newlength{\loclen}

\title{TMT: Cross-domain Semantic Segmentation with Region-adaptive Transferability Estimation}
\name{
Enming Zhang\textsuperscript{1} \enspace
Zhengyu Li\textsuperscript{2} \enspace
Yanru Wu\textsuperscript{1} \enspace
Jingge Wang\textsuperscript{1} \enspace
Yang Tan\textsuperscript{1} \enspace
Yang Li\textsuperscript{1}\sthanks{Corresponding author: yangli@sz.tsinghua.edu.cn} \enspace
Xiaoping Zhang\textsuperscript{1}
}
\address{
\textsuperscript{1}Tsinghua Shenzhen International Graduate School, Tsinghua University\sthanks{Shenzhen Key Laboratory of Ubiquitous Data Enabling} \\
\textsuperscript{2}Shanghai Jiao Tong University
}
\begin{document}
\ninept
\maketitle
\begin{abstract}

Recent advances in Vision Transformers (ViTs) have significantly advanced semantic segmentation performance. However, their adaptation to new target domains remains challenged by distribution shifts, which often disrupt global attention mechanisms. While existing global and patch-level adaptation methods offer some improvements, they overlook the spatially varying transferability inherent in different image regions. To address this, we propose the Transferable Mask Transformer (TMT), a region-adaptive framework designed to enhance cross-domain representation learning through transferability guidance.
First, we dynamically partition the image into coherent regions—grouped by structural and semantic similarity—and estimates their domain transferability at a localized level. Then, we incorporate region-level transferability maps directly into the self-attention mechanism of ViTs, allowing the model to adaptively focus attention on areas with lower transferability and higher semantic uncertainty.
Extensive experiments across 20 diverse cross-domain settings demonstrate that TMT not only mitigates the performance degradation typically associated with domain shift but also consistently outperforms existing approaches.
\end{abstract}
\begin{keywords}
Cross-Domain Adaptation, Transfer Learning, Transferability Estimation, Semantic Segmentation
\end{keywords}
\section{Introduction}
\begin{figure}[!ht]
  \centering
  \renewcommand\arraystretch{0}% 行间距
  \setlength\tabcolsep{1.7pt}% 列间距
  \begin{subfigure}{\linewidth}\centering%
    \begin{tabular}{ccc}%
      \includegraphics[height=40pt]{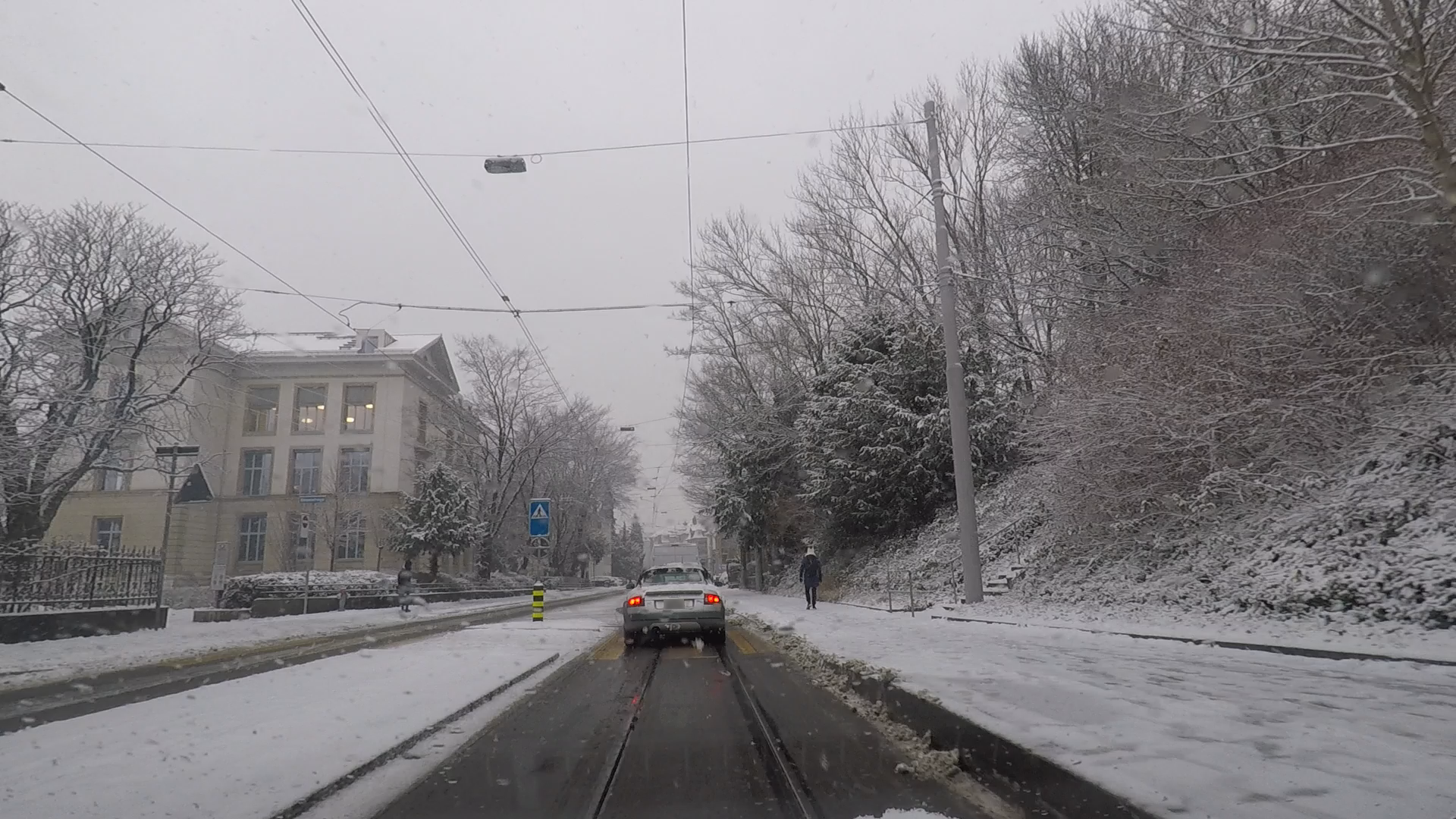}&%
      \includegraphics[height=40pt]{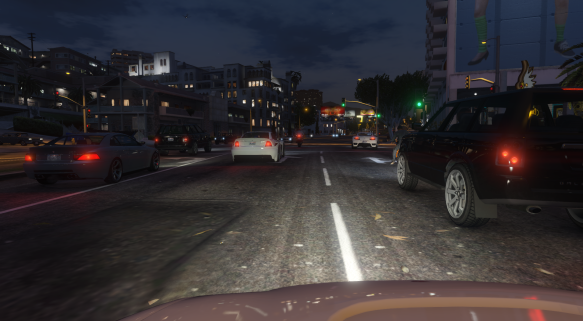}&%
      \includegraphics[height=40pt]{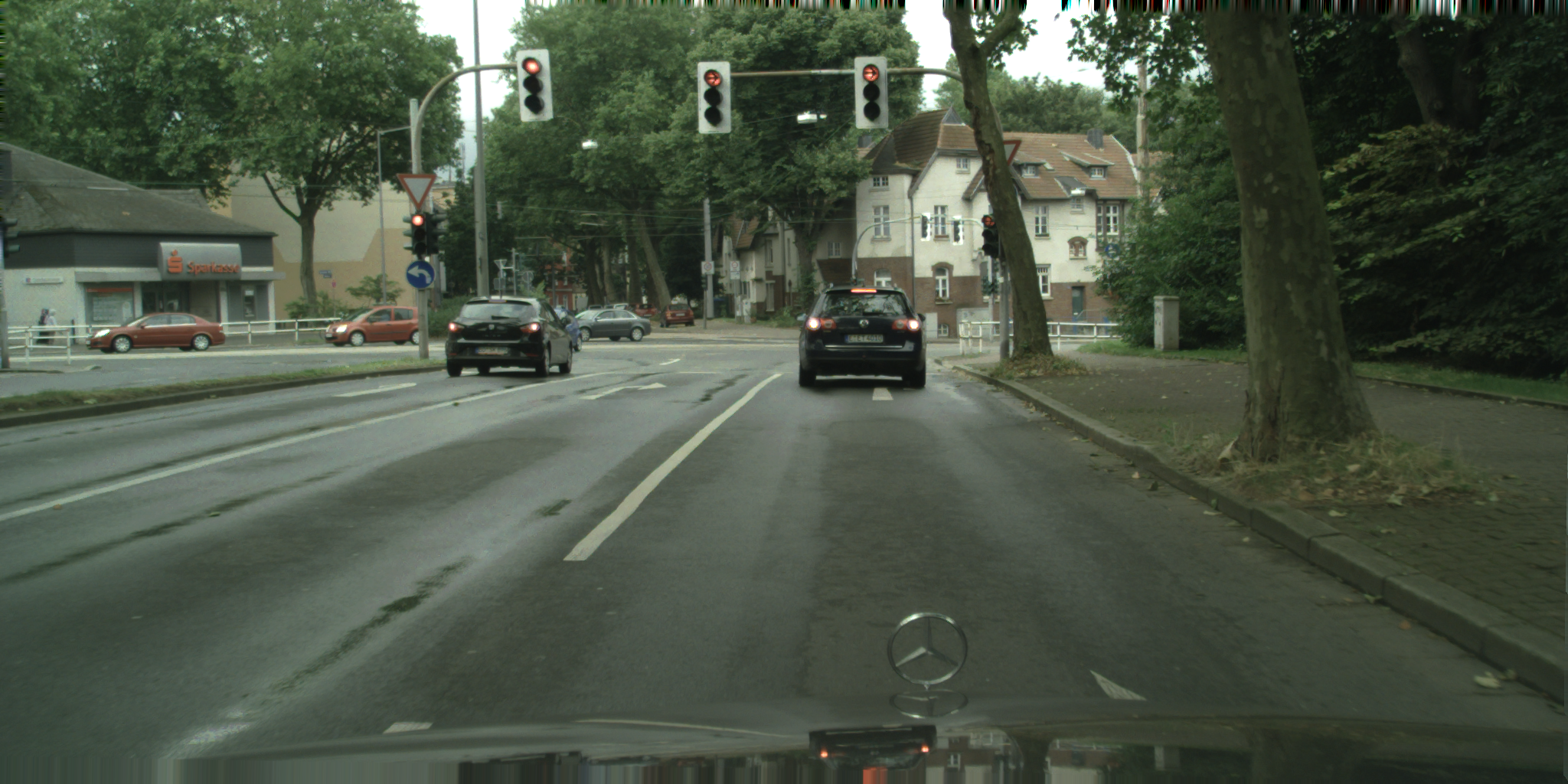}%
    \end{tabular}
    \caption{Examples of different datasets}
    \label{fig:3datasets}
  \end{subfigure}

  \begin{subfigure}{74.6pt}\centering%
    \begin{tabular}{c}\small
      \includegraphics[height=40pt]{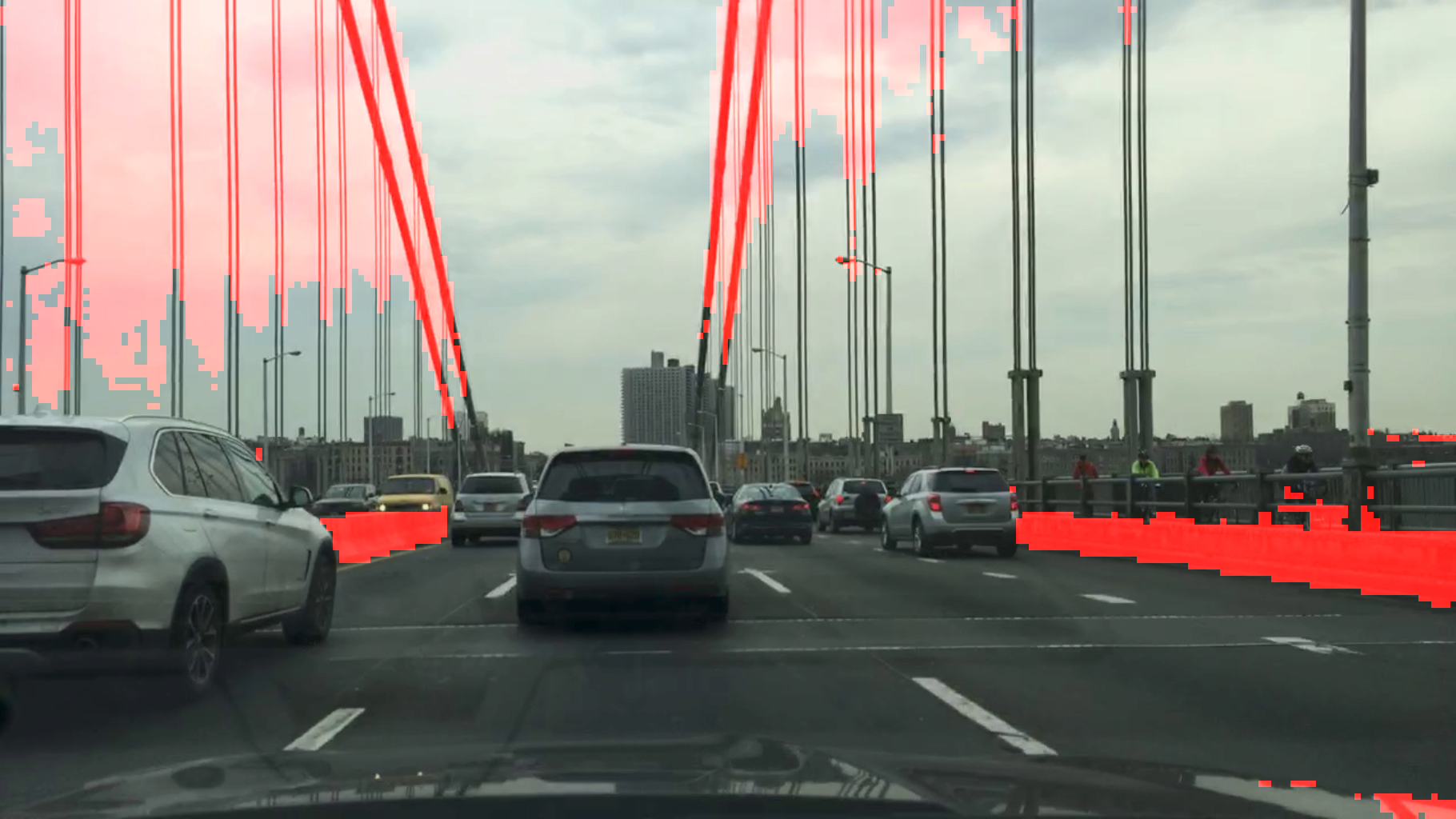}\\[0.5ex]
      \small Vanilla Finetuning \\[0.7ex]
      \includegraphics[height=40pt]{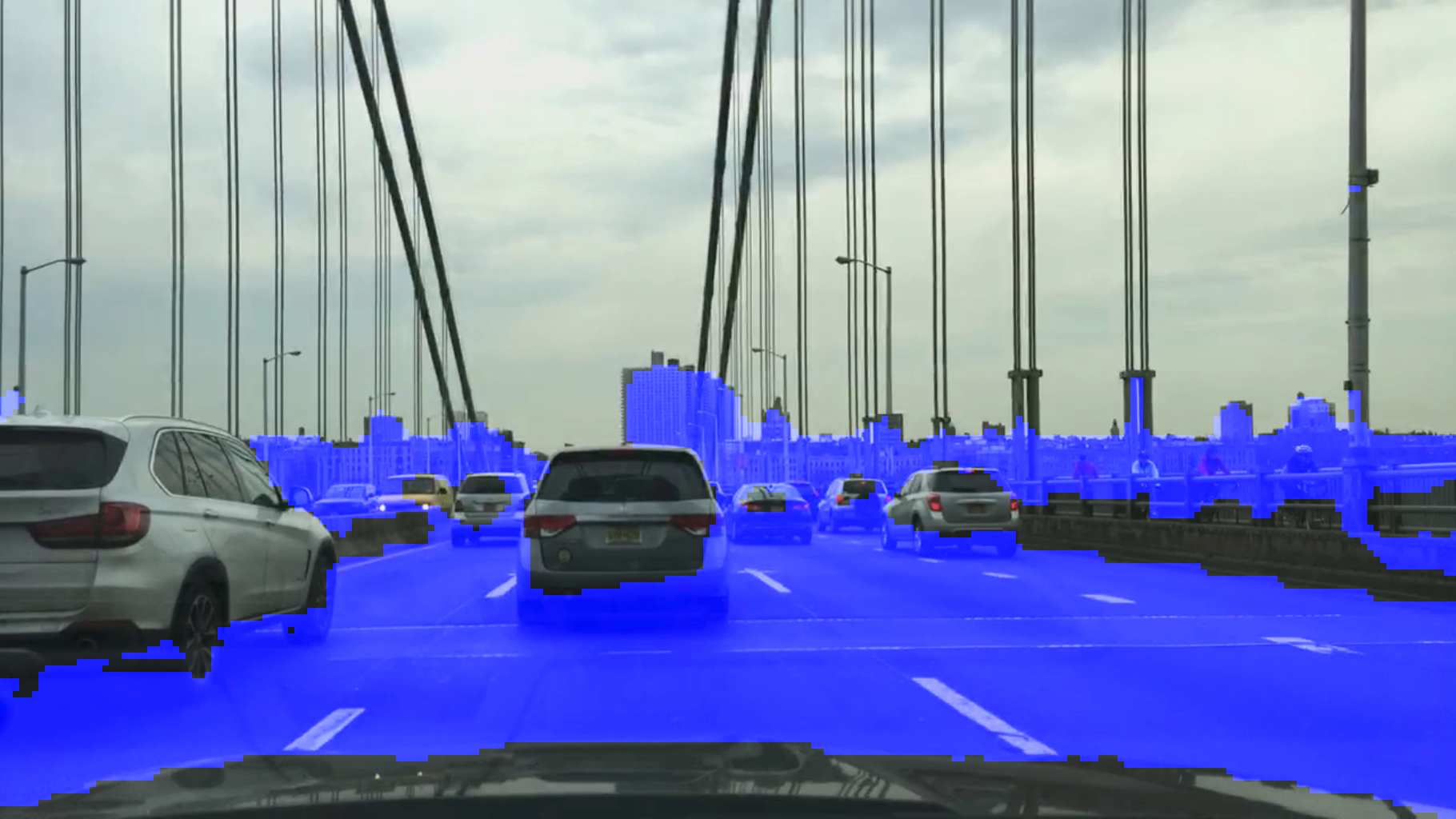}\\[0.5ex]
      \small \phantom{p}Ours\phantom{p}
    \end{tabular}
    \caption{Attention Mask}
    \label{fig:attentionmask}
  \end{subfigure}%
  \begin{subfigure}{74.6pt}\centering%
    \begin{tabular}{c}\small
      \includegraphics[height=40pt]{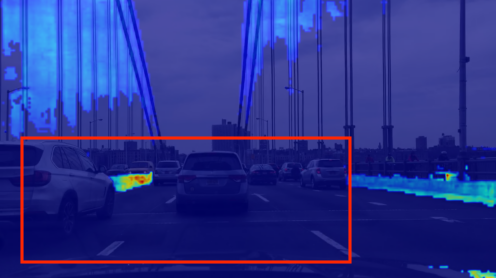}\\[0.5ex]
      \small Vanilla Finetuning \\[0.7ex]
      \includegraphics[height=40pt]{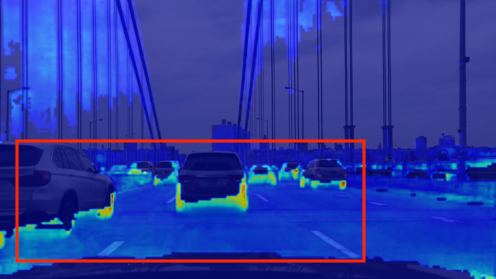}\\[0.5ex]
      \small \phantom{p}Ours\phantom{p}
    \end{tabular}
    \caption{Attention Map}
    \label{fig:attentionmap}
  \end{subfigure}%
  \begin{subfigure}{84.2pt}\centering%
    \begin{tabular}{c}
      \includegraphics[height=40pt]{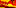}\\[0.5ex]
      \small Equal-sized Patch\\[0.7ex]
      \includegraphics[height=40pt]{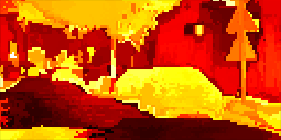}\\[0.5ex]
      \small Adaptive Region%
    \end{tabular}
    \caption{Partition Method}
    \label{fig:切分}
  \end{subfigure}%
  \caption{\textbf{Fig.a} illustrates domain gaps across datasets, reflecting real-world challenges. \textbf{Fig.b–c} show that vanilla fine-tuning produces misleading attention masks (red), focusing on irrelevant areas, while our method yields accurate masks (blue) by emphasizing task-critical regions like the car. \textbf{Fig.d} compares region partition methods, with our approach adaptively segmenting images for better transferability estimation.} 
    \vspace{-0.3cm}
\end{figure}

\begin{figure*}[t!]
  \centering
  \includegraphics[width=0.9\textwidth]{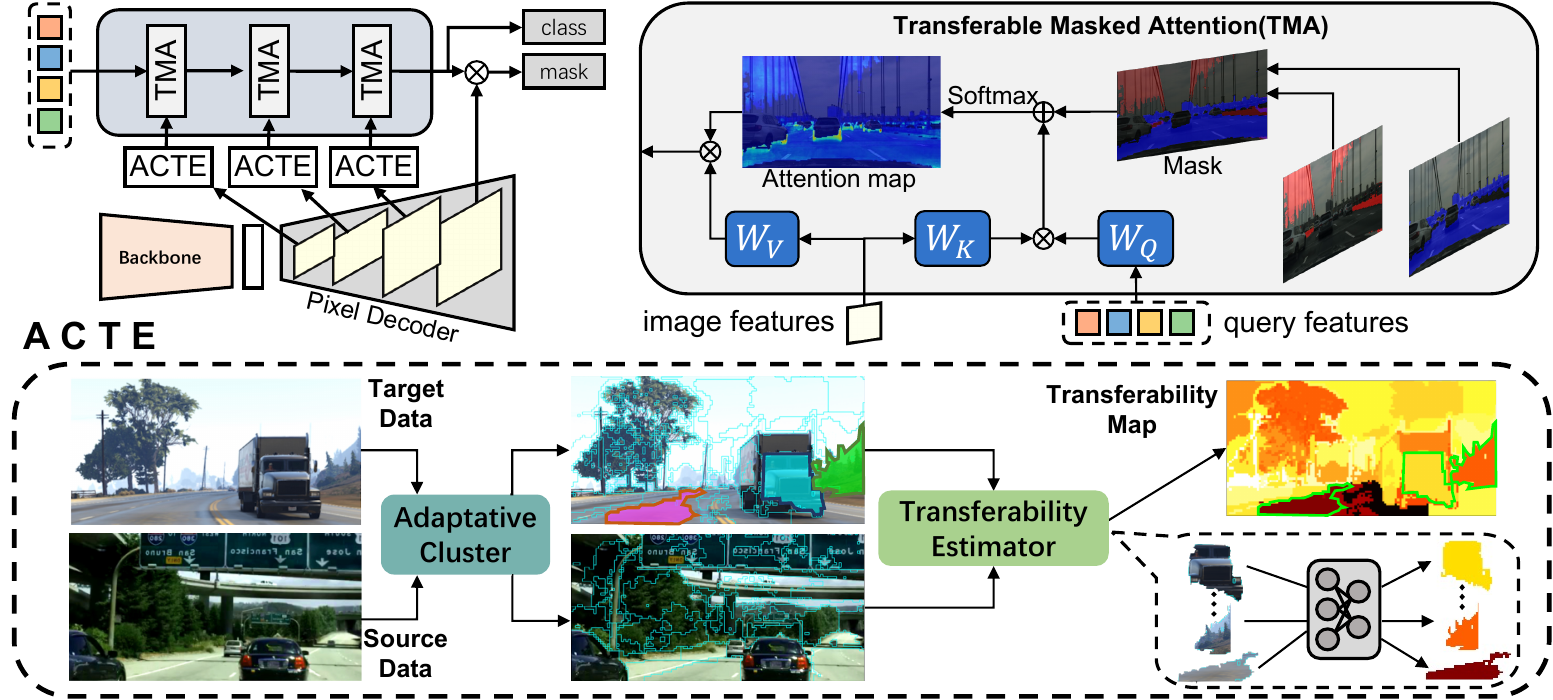} 
  \caption{Overview of the framework. The model training begins with the ACTE, which is first trained using both source and target data. The lower section illustrates how ACTE evaluates and assigns different transferability scores to various regions, represented by different colors in the transferability map. After ACTE has been trained, its output—region-level transferability maps—is used to guide the training of the TMA within attention mechanism (top-right).}
  \label{fig:主图}
    \vspace{-0.4cm}
\end{figure*}

Semantic segmentation, a core task in computer vision, assigns semantic categories to each pixel in an image. While vision transformers \cite{he2017mask, xiao2018unified} have surpassed CNN-based models and achieved state-of-the-art performance, fine-tuning these models remains challenging due to distribution shifts\cite{kirillov2023segany,wang2024implicitexplicitlanguageguidance,huang2023stunetscalabletransferablemedical,guo2019spottune,liu2024luminamgptilluminateflexiblephotorealistic}. As shown in Fig. \ref{fig:3datasets}, domain gaps arise when source data differs significantly from the target domain due to variations in visual features like lighting, background, or object appearances. These shifts are particularly problematic for vision transformers, whose global attention mechanisms can be easily disrupted. For instance, Fig. \ref{fig:attentionmask} and \ref{fig:attentionmap} demonstrates that vanilla fine-tuning often leads to misdirected attention maps, where the model fails to attend to semantically related patches. This phenomenon aligns with previous observations \cite{wang_cdac_2023} that distribution shifts can degrade model performance due to difference in object co-occurrence patterns. Therefore, developing a more fine-grained fine-tuning method is essential to address these challenges and achieve robust performance on unseen and domain-diverse segmentation tasks.

To address the challenges of distribution shifts in vision transformers, several approaches have been proposed. For instance, methods like \cite{wang_cdac_2023} and \cite{kirillov2023segany} focus on aligning feature distributions between source and target domains, while others \cite{guo2019spottune, liu2024luminamgptilluminateflexiblephotorealistic} employ more advanced fine-tuning strategies to mitigate domain gaps. However, these methods often treat the entire image as a uniform unit, not taking into account the spatial heterogeneity in transferability across regions in images from different domains \cite{wang_transferable_2019, xiao_not_2023, chen_sparsevit_2023}. For example, in autonomous driving scenarios, background regions like the sky are highly consistent across domains and thus easily transferable, while intricate urban elements pose greater challenges with lower transferability due to their complexity and variability. This discrepancy suggests that a global or patch-level approach to fine-tuning is insufficient, as it overlooks the need for region-adaptive transferability assessment.

To address this limitation, we propose a novel Adaptive Cluster-based Transferability Estimator (ACTE), which flexibly evaluates region-level transferability. While global and patch-level domain adaptation methods offer partial solutions \cite{wang_transferable_2019, yang_tvt_2023, tan_efficient_2023}, they often fall short in capturing regional semantics and tend to disrupt structural and semantic consistency, especially at object boundaries or for small objects, as demonstrated in Fig. \ref{fig:切分}. In contrast, our method introduces a hierarchical and adaptive approach that dynamically segments images into regions based on semantic and structural coherence. By preserving object boundaries and handling small objects effectively, ACTE ensures a localized assessment of transferability, therefore significantly improving cross-domain generalization.

While existing methods for leveraging transferability have demonstrated success in CNN-based architectures \cite{tan_efficient_2023}, their direct application to vision transformers remains underexplored. To bridge this gap, we propose Transferable Masked Attention (TMA) tailored for transformer-based models. TMA dynamically adjusts attention masks by incorporating region-specific transferability maps, enabling object queries to focus on regions with both low transferability and low semantic confidence. By combining transferability scores in a hierarchical way, TMA ensures that the model prioritizes regions requiring adaptation while suppressing attention to well-aligned or confidently predicted areas. This approach not only mitigates inappropriate mask predictions but also enhances the model’s ability to capture domain-specific details, offering a principled way to integrate transferability estimation into the attention mechanism of vision transformers.

Our experimental results show that the proposed methods achieve an average MIoU improvement of 2.1\% compared to vanilla fine-tuning and 1.28\% over the SOTA method across 20 source-target transfer pairs over 5 popular semantic segmentation benchmarks for autonomous driving.

\vspace{-1em}
\section{Method}
Our framework, TMT, illustrated in Fig. \ref{fig:主图}, builds upon the Mask2Former architecture to address cross-domain image segmentation through a unified and layer-wise adaptive pipeline. After extracting hierarchical feature maps, the model adaptively segments the image into regions based on structural characteristics. A Transferability Estimator is then employed to assess region-wise transferability. This transferability map is integrated into every layer of the network, dynamically modulating the attention mechanism to enhance domain adaptation across multiple feature levels.
\vspace{-1em}
\subsection{Adaptive Cluster}

Our adaptive part-level grouping method generates structurally coherent regions aligned with image boundaries while preserving semantic consistency within each region. Following the common practice in superpixel learning, we employ an iterative grouping algorithm that processes feature maps to produce part-level mask features. The pixel space is initially divided into a regular grid of size $r \times r$, where each cell serves as an initial cluster center, yielding $N_p = \frac{H}{r} \times \frac{W}{r}$ centers. During each iteration, we first compute the cluster center features $\mathbf{Q} \in \mathbb{R}^{N_p \times d}$ by averaging the pixel features within each region. Next, we compute the cosine similarity between each pixel and its surrounding $3 \times 3$ centers. The similarity between the $i$-th center feature $\mathbf{Q}_i$ and the $j$-th pixel feature $\mathbf{K}_j$ is given by:
\begin{equation}
D_{i,j} = 
\begin{cases} 
f(\mathbf{Q}_i, \mathbf{K}_j) = \dfrac{1}{\tau} \cdot \dfrac{\mathbf{Q}_i \cdot \mathbf{K}_j}{\|\mathbf{Q}_i\| \|\mathbf{K}_j\|} & \text{if } i \in \mathcal{N}_j \\
-\infty & \text{if } i \notin \mathcal{N}_j
\end{cases}
\end{equation}
where $\mathcal{N}_j$ denotes the set of nearby centers of the $j$-th pixel, and $\tau$ is a temperature parameter.

The soft assignment matrix $\mathbf{A} \in [0,1]^{N_p \times (HW)}$ is then derived by applying softmax normalization over the similarity values for each pixel:
\begin{equation}
\mathbf{A}_{i,j} = \frac{\exp(D_{i,j})}{\sum_{i=1}^{N_p} \exp(D_{i,j})}
\end{equation}
The cluster center features are updated via:
\begin{equation}
\mathbf{Q}^{\text{new}} = \mathbf{A} \times \mathbf{K}
\end{equation}
This iterative refinement process converges to a final clustered feature map where each region corresponds to a coherent image segment, effectively reducing computational complexity through local constraints while maintaining structural alignment with image content.

\vspace{-1em}
\subsection{Region-level Transferability Estimator}
To quantify the domain shift between source and target domains, we adopt the \textit{Proxy $\mathcal{A}$-distance (PAD)}~\cite{ganin2016domain}, which measures the divergence between the marginal distributions $\mathcal{D}_{\mathrm S}^X$ and $\mathcal{D}_{\mathrm T}^X$. The $\mathcal{A}$-divergence is defined as:
\begin{equation}
  d_{\mathcal A}(\mathcal D_{\mathrm S}^X,\mathcal D_{\mathrm T}^X):=2\sup_{A\in \mathcal A} \left|\Pr[\bm x_{\mathrm S}\in A]-\Pr[\bm x_{\mathrm T}\in A]\right|.
\end{equation}
In practice, we estimate this divergence by training a domain discriminator $E$ to classify samples from the source and target domains. The Proxy $\mathcal{A}$-distance is then approximated based on the generalization error of this discriminator. We extend this PAD framework to operate at the region level rather than at the image level, enabling a fine-grained assessment of transferability across different regions within each image.

We design a domain discriminator $E$ that takes region-level features $\mathbf{Q}_i$ from both domains as input and predicts the probability that the region originates from the target domain. The discriminator is optimized using a cross-entropy loss:
\begin{equation}
  \mathcal L(E(\mathbf{Q}_i),d)
  =(1-d)\log\frac{1}{1 - E(\mathbf{Q}_i)} + d\log\frac{1}{E(\mathbf{Q}_i)},
\end{equation}
where $d$ denotes the domain label, with $d = 1$ for source domains and $d = 0$ for the target domain. To train \(E\), we construct a balanced training set comprising region features from both domains, each annotated with their corresponding domain labels. Batches are uniformly sampled from both domains to ensure balanced supervision during training.

Regions with high \(\bm T_{i}\) require little adaptation, as they already exhibit strong transferability and offer limited potential for further improvement. In contrast, regions with lower \(\bm T_{i}\) exhibit substantial distribution shifts relative to the source domain and therefore require more targeted adaptation to reduce the domain gap. By focusing on these low-transferability regions, we can more effectively adapt the model to the target domain’s characteristics.
 
\vspace{-1em}
\subsection{Transferability Guided Masked Attention}
With the region-level transferability map \(\bm T\) obtained from the domain discriminator \(E\), we now leverage this critical information to guide the transformer's learning process. The transferability map \(\bm T\) encodes the degree of domain alignment for each region, providing a principled way to modulate the attention mechanism in vision transformers. Building upon Mask2Former's architecture \cite{cheng_masked-attention_2022}, we introduce a key innovation: the integration of region-aware transferability maps into the masked attention mechanism, enabling dynamic adaptation to domain shifts.  

Formally, the Transferable Masked Attention (TMA) operation is formulated as:  
\begin{equation}
  \operatorname{Softmax}\left(\bm{\mathcal{M}}(\bm T) + \frac{\bm{K}^{\mathrm{T}}\bm{Q}}{\sqrt{C}}\right) \cdot \bm{V},
\end{equation}  
where \(\bm{\mathcal{M}}(\bm T)\) is a transferability-aware attention mask dynamically conditioned on the transferability map \(\bm T\). Here, \(\bm{K} \in \mathbb{R}^{C \times H_lW_l}\) and \(\bm{Q} \in \mathbb{R}^{C \times N}\) are the key and query feature matrices, respectively, and \(\bm{V} \in \mathbb{R}^{H_lW_l \times C}\) is the value matrix. The mask \(\bm{\mathcal{M}}(\bm T)\) is defined as:  
\begin{equation}
  \bm{\mathcal{M}}_{i,j}(\bm T) = 
  \begin{cases} 
    0, & \text{if } \bm{M}_{i,j} \leq \lambda_M \text{ and } \bm{T}_{i,j} \leq \lambda_{\mathrm{T}}, \\
    -\infty, & \text{otherwise},
  \end{cases}
\end{equation}  
where \(\bm{M}_{i,j}\) and \(\bm{T}_{i,j}\) represent the predicted attention mask from the transformer decoder and the transferability scores, respectively. TMA dynamically adjusts the attention mechanism based on both semantic uncertainty and domain divergence, focusing on regions with low semantic confidence (\(\bm{M}_{i,j} \leq \lambda_M\)) and low transferability (\(\bm{T}_{i,j} \leq \lambda_{\mathrm{T}}\)). The thresholds \(\lambda_M\) and \(\lambda_{\mathrm{T}}\) control the strictness of this selection process, as analyzed in Section \ref{Parameter_Analysis}. This dual-conditioned masking ensures that the model prioritizes regions where adaptation is most critical, while suppressing attention to well-predicted and domain-aligned areas.

\begin{table*}[t!]
\centering
\renewcommand\arraystretch{1.1}
\setlength\tabcolsep{0.5pt}
\setlength{\loclen}{2.3em}
\begin{tabular}{c|c|C{\loclen}C{\loclen}C{\loclen}C{\loclen}|C{\loclen}C{\loclen}C{\loclen}C{\loclen}|C{\loclen}C{\loclen}C{\loclen}C{\loclen}|C{\loclen}C{\loclen}C{\loclen}C{\loclen}|C{\loclen}C{\loclen}C{\loclen}C{\loclen}|C{\loclen}}
\Xhline{1.2pt}
\multirow{2}{*}{} & \multirow{2}{*}{\textbf{Method}} 
& \multicolumn{4}{c|}{C} 
& \multicolumn{4}{c|}{B} 
& \multicolumn{4}{c|}{M} 
& \multicolumn{4}{c|}{S} 
& \multicolumn{4}{c|}{G} 
& \multirow{2}{*}{\textbf{Avg}} \\\cline{3-22}
\hhline{~|~|----|----|----|----|----|~}
& & B & M & S & G & C & M & S & G & C & B & S & G & C & B & M & G & C & B & M & S & \\\Xhline{1.2pt}

\multirow{7}{*}{\rotatebox[origin=c]{90}{MIoU}} 
& Linear & 55.0 & 67.3 & 43.8 & 56.7 & 47.3 & 47.0 & 29.3 & 39.8 & 72.4 & 61.9 & 48.8 & 64.4 & 31.1 & 22.7 & 28.7 & 28.4 & 38.3 & 30.2 & 40.9 & 27.8 & 44.1 \\
& DAF & 57.8 & 67.8 & 52.5 & 59.9 & 50.0 & 46.9 & 36.2 & 41.1 & 72.8 & 62.1 & 53.6 & 65.2 & 36.0 & 26.9 & 34.4 & 33.4 & 41.8 & 32.7 & 43.6 & 39.5 & 47.7 \\ 
& MIC & 60.6 & 69.8 & 66.3 & 66.3 & 53.1 & 47.0 & 47.3 & 47.8 & \underline{73.0} & 62.6 & 60.1 & 70.6 & 45.4 & 31.7 & 38.2 & 43.1 & 51.6 & 38.7 & 45.9 & 48.0 & 53.4 \\
& Full & 62.2 & 70.7 & 75.8 & 69.2 & 54.7 & 46.9 & 53.0 & 52.0 & \textbf{73.6} & 61.2 & 67.3 & 71.7 & 50.9 & 37.6 & 43.7 & 49.7 & 57.0 & 42.3 & 48.4 & 57.2 & 57.3 \\
& LEEP & 62.8 & 71.1 & 76.2 & 69.5 & 55.2 & 47.3 & 53.4 & 52.5 & 72.6 & \underline{62.7} & 68.5 & \underline{72.5} & 51.2 & 38.0 & \underline{44.0} & 50.0 & 57.5 & 42.8 & 48.8 & 57.6 & 57.7 \\
& OTCE & \underline{63.0} & \underline{71.4} & \underline{76.5} & \underline{69.3} & \underline{55.5} & \underline{47.6} & \underline{53.7} & \underline{52.8} & 72.8 & 62.3 & \textbf{69.8} & 71.8 & \underline{51.5} & \underline{38.2} & \textbf{44.3} & \underline{50.3} & \underline{57.8} & \underline{43.0} & \underline{49.0} & \underline{57.9} & \underline{57.9} \\
& \cellcolor{gray!20} 
\textbf{Ours} & \cellcolor{gray!20} \textbf{64.5} & \cellcolor{gray!20} \textbf{72.7} & \cellcolor{gray!20} \textbf{77.8} & \cellcolor{gray!20} \textbf{71.2} & \cellcolor{gray!20} \textbf{57.5} & \cellcolor{gray!20} \textbf{49.0} & \cellcolor{gray!20} \textbf{55.8} & \cellcolor{gray!20} \textbf{54.8} & \cellcolor{gray!20} \underline{73.0} & \cellcolor{gray!20} \textbf{62.9} & \cellcolor{gray!20} \underline{69.5} & \cellcolor{gray!20} \textbf{73.5} & \cellcolor{gray!20} \textbf{53.5} & \cellcolor{gray!20} \textbf{41.1} & \cellcolor{gray!20} \underline{44.0} & \cellcolor{gray!20} \textbf{51.8} & \cellcolor{gray!20} \textbf{59.5} & \cellcolor{gray!20} \textbf{44.8} & \cellcolor{gray!20} \textbf{51.0} & \cellcolor{gray!20} \textbf{59.5} & \cellcolor{gray!20} \textbf{59.4} \\\Xhline{1.2pt}

\multirow{7}{*}{\rotatebox[origin=c]{90}{mACC}} 
& Linear & 68.4 & 79.9 & 64.1 & 70.4 & 58.4 & 58.1 & 36.6 & 50.7 & 83.9 & \underline{73.5} & 68.2 & 78.1 & 38.9 & 28.3 & 36.2 & 39.5 & 47.6 & 36.5 & 51.5 & 40.1 & 55.4 \\
& DAF & 69.6 & 80.2 & 67.5 & 71.6 & 59.7 & 57.2 & 39.5 & 53.8 & 84.2 & 73.2 & 74.3 & 78.9 & 45.7 & 31.6 & 39.7 & 46.0 & 52.8 & 42.0 & 53.4 & 49.3 & 58.5 \\
& MIC & 73.1 & 80.5 & 80.8 & 78.3 & 64.4 & 57.9 & 58.9 & 58.2 & 84.6 & \underline{73.5} & 81.4 & 80.6 & 55.2 & 39.8 & 49.6 & 54.4 & 60.4 & 47.7 & 58.6 & 58.5 & 64.8 \\
& Full & 73.8 & 80.7 & 85.2 & 79.9 & 67.5 & 57.1 & 63.4 & 61.8 & \underline{84.9} & 73.1 & 84.9 & 81.7 & 62.5 & 46.0 & 54.0 & 60.2 & 68.9 & 51.2 & 59.7 & 67.7 & 68.2 \\
& LEEP & 74.2 & 81.1 & 85.5 & 80.2 & 67.9 & 57.5 & 63.8 & 62.2 & 84.6 & 72.8 & 85.9 & 82.5 & 62.9 & 46.4 & 54.3 & 60.5 & 69.2 & 51.7 & 60.0 & 68.0 & 68.6 \\
& OTCE & \underline{74.5} & \underline{81.4} & \underline{85.8} & \underline{80.5} & \underline{68.2} & \underline{57.8} & \underline{64.0} & \underline{62.5} & 84.3 & 72.5 & \textbf{87.3} & \textbf{83.9} & \underline{63.2} & \underline{46.7} & \underline{54.6} & \underline{60.8} & \underline{69.5} & \underline{52.0} & \underline{60.3} & \underline{68.1} & \underline{68.9} \\
& \cellcolor{gray!20} \textbf{Ours} & \cellcolor{gray!20} \textbf{75.8} & \cellcolor{gray!20} \textbf{82.8} & \cellcolor{gray!20} \textbf{87.1} & \cellcolor{gray!20} \textbf{81.7} & \cellcolor{gray!20} \textbf{70.2} & \cellcolor{gray!20} \textbf{59.8} & \cellcolor{gray!20} \textbf{66.4} & \cellcolor{gray!20} \textbf{64.3} & \cellcolor{gray!20} \textbf{85.2} & \cellcolor{gray!20} \textbf{73.7} & \cellcolor{gray!20} \underline{87.0} & \cellcolor{gray!20} \underline{83.7} & \cellcolor{gray!20} \textbf{65.7} & \cellcolor{gray!20} \textbf{48.5} & \cellcolor{gray!20} \textbf{56.1} & \cellcolor{gray!20} \textbf{62.2} & \cellcolor{gray!20} \textbf{71.0} & \cellcolor{gray!20} \textbf{54.0} & \cellcolor{gray!20} \textbf{62.0} & \cellcolor{gray!20} \textbf{69.8} & \cellcolor{gray!20} \textbf{70.4} \\\Xhline{1.2pt}
\end{tabular}

\caption{Performance comparison (MIoU and mACC) of different methods across various domain adaptation scenarios.}
\label{table：主表}
\vspace{-0.3cm}
\end{table*}

\section{Experiments}
\subsection{Datasets}
% \textbf{Cityscapes} dataset \cite{Cordts2016Cityscapes} includes 5000 images of 2048×1024 pixels from 50 German cities, diverse in seasons, times, backgrounds and weather. \textbf{BDD} dataset \cite{bdd100k} contains 7000 training and 1000 testing images of 1280×720 pixels, collected from US street scenes. \textbf{Mapillary} \cite{Mapillary} is a large-scale dataset with 18000 training images, 2000 validation images, and 5000 testing images from around the world, showcasing high diversity in weather and seasonal conditions. For synthetic datasets, \textbf{GTAV} \cite{Richter_2016_ECCV} contains highly realistic images generated from \textit{Grand Theft Auto V} at 1914×1052 pixels, with 12,403 training, 6,382 validation, and 6,181 testing images. Additionally, \textbf{SYNTHIA} \cite{Ros_2016_CVPR} has 9400 images of 1280×760 pixels.

In our experiments, we utilized five different source datasets — Cityscapes \cite{Cordts2016Cityscapes}, BDD \cite{bdd100k}, Mapillary \cite{Mapillary}, SYNTHIA \cite{Ros_2016_CVPR}, and GTAV \cite{Richter_2016_ECCV}—and evaluated the segmentation performance of models to various target datasets. Each dataset represents different challenges: Cityscapes and BDD are real-world urban datasets, Mapillary offers diverse scenes, SYNTHIA and GTAV are synthetic datasets with varying degrees of realism. The purpose of using such a diverse set of datasets is to assess how well the models can generalize across different domains, particularly from synthetic to real-world scenarios.
\vspace{-1em}
\subsection{Training Details}
To achieve true transfer accuracy in each transfer experiment, we train the source model for a total of 20,000 iterations, evaluating the test accuracy every 1,000 steps. During the transfer training phase, the estimator is first trained and then frozen to stabilize its evaluation. We then initialize the target model with the pre-trained weights from the source model and proceed to fine-tune the entire model using the AdamW optimizer, with a learning rate of 0.0001 and a batch size of 16. The model training is conducted using two NVIDIA A800 GPUs. The cluster algorithm repeats its steps for 6 iterations.
\vspace{-1em}

\subsection{Baselines}
\begin{figure}[t!]
  \centering
  \includegraphics[width=0.45\textwidth]{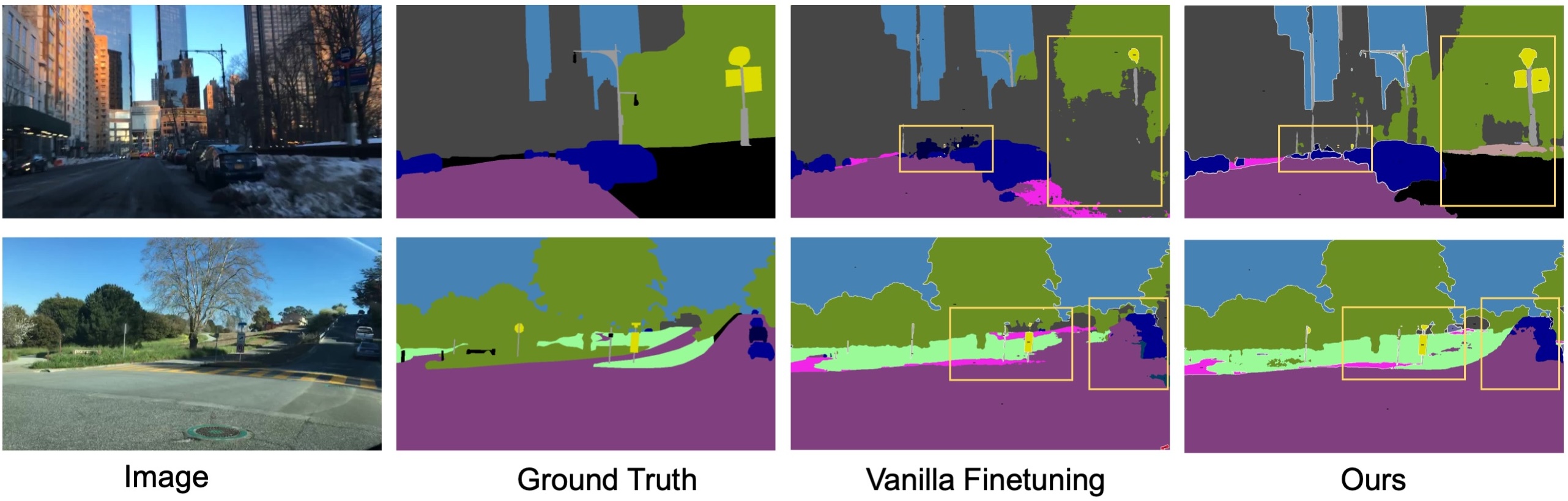}
  \caption{Visualization of segmentation results where models pretrained on Cityscapes are transferred to the BDD dataset.}
  \label{fig:分割结果图}
\end{figure}

In our experimental design, we include classic adaptation strategies such as partial network adaptation via linear probing and full-parameter fine-tuning. We also incorporate the seminal DAFormer architecture \cite{hoyer2022hrda}, which is specifically designed for robust semantic segmentation under unsupervised domain adaptation (UDA), and further enhance it with the context-aware self-training method MIC \cite{hoyer2023mic} to evaluate the performance gains from improved contextual reasoning. Additionally, we compare our approach with two state-of-the-art transferability-aware methods most closely related to ours: OTCE-Finetuning and LEEP-Finetuning \cite{tan2024transferability}.

\vspace{-1.5em}
\subsection{Main Results}
Our proposed method consistently shows superior performance across the majority of source-target pairs when compared to baseline in Tab.\ref{table：主表}. The MIoU across the 20 source-target pairs in our experiments shows an average increase of 2\% compared to full fine-tuning and an average increase of 1.28\% compared to the OTCE-finetuning method.

When Cityscapes or BDD is used as the source, our method consistently achieves the highest scores across all target datasets. This demonstrates the strong generalization ability of our approach when transferring from well-curated, real-world datasets. SYNTHIA and GTAV, being synthetic datasets, both showcase our method's robust adaptability and mitigate the domain gap, particularly in handling synthetic-to-real domain shifts, which are typically challenging.

As shown in Fig.\ref{fig:分割结果图}, our method more accurately delineates the boundaries of vehicles and roads, preserving the complete structure of objects. It also demonstrates a superior ability in segmenting small objects.

\begin{figure}[t!]
\centering
\renewcommand\arraystretch{1}% 行间距
\setlength\tabcolsep{0.16em}% 列间距
% 计算新宽度（每行4张图）
\newlength{\newloclen}
\setlength{\newloclen}{0.24\linewidth} % 约1/4宽度
\addtolength{\newloclen}{-0.24em} % 微调间距

\begin{tabular}{cccc}
\includegraphics[width=\newloclen]{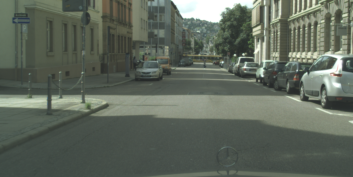} & 
\includegraphics[width=\newloclen]{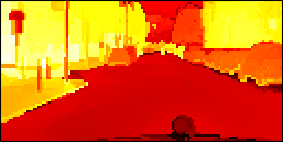} & 
\includegraphics[width=\newloclen]{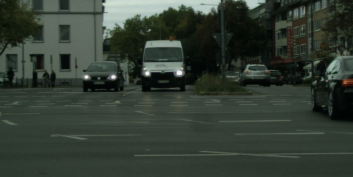} & 
\includegraphics[width=\newloclen]{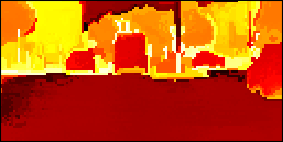} \\
\includegraphics[width=\newloclen]{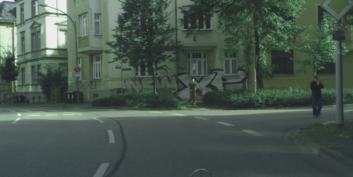} & 
\includegraphics[width=\newloclen]{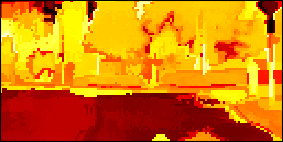} & 
\includegraphics[width=\newloclen]{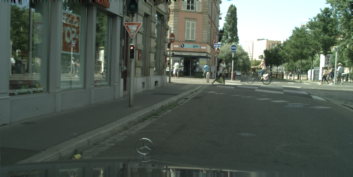} & 
\includegraphics[width=\newloclen]{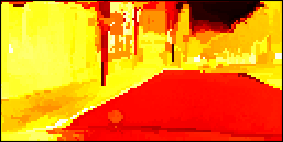}
\end{tabular}

\caption{Visualization of transferability maps between the Cityscapes (target domain) and Mapillary (source domain) datasets. The maps indicate regions with high transferability in deep colors and low transferability in lighter colors.}
\label{fig:transferability_map}
\end{figure}

\vspace{-1.5em}
\subsection{Transferability Maps}
The transferability maps generated from the experiments clearly demonstrate the effectiveness of our method. Each pair of images in Fig.\ref{fig:transferability_map} shows the original input on the left and the corresponding transferability map on the right. The target domain is Cityscapes, and the source domain is Mapillary, both of which consist of real urban scenes. Therefore, elements such as vehicles are relatively similar across both domains. This observation is consistent with real-world scenarios, as the features of roads and skies are relatively simple and consistent across different domains, while the features of buildings are more complex and variable. This outcome reflects that the domain discriminator effectively identifies the complex and diverse image information present in urban environments and accurately assesses the transferability of different regions.

% \subsubsection{Attention Mechanism}
% We aim to provide a detailed analysis of the internal mechanisms driving the effectiveness of our method in Fig \ref{fig:attention}. The first column highlights red regions where the vanilla finetuning misdirects masks to less relevant areas, such as the sky, leading to suboptimal segmentation, particularly in tasks like autonomous driving where road layout understanding is critical. In contrast, the second column shows our methods segmentation masks, with blue regions indicating a more accurate focus on important areas. The fourth and fifth columns further demonstrate how our method directs attention to crucial areas, such as the road, rather than irrelevant regions.

\vspace{-1.5em}
\subsection{Ablation Study}
\begin{table}[tb!]\centering
  \def\→{\raisebox{0.3ex}{\scriptsize\textbf{→}}}
  \renewcommand\arraystretch{1.12}% 行间距
  \setlength\tabcolsep{0.139em}% 列间距
  \begin{tabular}{ccccccccc}
  \Xhline{0.75pt}
  & C\→B & C\→M & C\→S & C\→G & B\→C & B\→M & B\→S & B\→G\!\\
  \Xhline{0.75pt}\normalsize
  \!w/o ACTE & 63.5 & 71.7 & 76.8 & 70.2 & 56.5 & 48.0 & 54.8 & 53.8 \\
  w/o TMA & 63.0 & 71.2 & 76.3 & 69.7 & 56.0 & 47.5 & 54.3 & 53.3 \\
  Ours & \textbf{64.5} & \textbf{72.7} & \textbf{77.8} & \textbf{71.2} & \textbf{57.5} & \textbf{49.0} & \textbf{55.8} & \textbf{54.8} \\\Xhline{0.75pt}
  \end{tabular}
  \caption{Ablation Study Results.}
  \label{table:ablation}
  \vspace{-0.3cm}
\end{table}

We conducted an ablation study on the transfer from Cityscapes and BDD to other datasets, as summarized in Tab. \ref{table:ablation}. Removing ACTE leads to a slight performance drop of approximately 1\%in mIoU, indicating that its adaptive clustering helps align features with target data structures. When TMA is removed, the transferability map is no longer incorporated into the attention mechanism. Instead, it is directly integrated into the final loss function via a weighted scheme similar to OTCE-finetuning, resulting in a more significant decline of around 1.5\%. This underscores the importance of mask-guided attention in refining cross-domain feature extraction. The full model combining both ACTE and TMA achieves the best performance, demonstrating that the two components are complementary.
\vspace{-1em}
\subsection{Parameter Analysis}
\label{Parameter_Analysis}
% \begin{figure}[t!]
%   \centering
%   \includegraphics[width=0.99\linewidth]{ResultImage/parameter.jpg}
%   \caption{MIoU changes as \(p_{\mathrm T}\) varies from 10\% to 50\%.}
%   \label{fig:param}
%   \vspace{-0.3cm}
% \end{figure}
The parameter $p_T$ determines the percentile threshold $\lambda_T$ used to identify transferable regions in the attention mechanism. Performance consistently improves as $p_T$ increases from $10\%$ to $30\%$, peaking at $30\%$. Beyond this value, performance declines across all target domains. Based on these results, we set $p_T = 30\%$ as the optimal value.

\vspace{-1em}
\section{Conclusion and Future work}

This paper presents a novel framework for semantic segmentation transfer learning, leveraging the Adaptive Cluster-based Transferability Estimator (ACTE) and Transferable Masked Attention (TMA) to address the challenges of domain adaptation in vision transformers. ACTE dynamically evaluates region-level transferability, enabling targeted adaptation by focusing on the most informative and domain-divergent regions. TMA integrates transferability maps into Mask2Former's attention mechanism, enhancing the model's ability to prioritize regions with low transferability and semantic uncertainty. Extensive experiments across multiple benchmarks demonstrate the effectiveness of our approach, achieving significant improvements over both vanilla fine-tuning and state-of-the-art methods. Looking ahead, future work could investigate cross-modal adaptability and integrate emerging architectural innovations to further advance the field. These directions hold the potential to develop more intelligent and adaptable visual systems, capable of handling the complexities of real-world scenarios with greater efficiency and accuracy.

\bibliographystyle{IEEEbib}
\bibliography{strings,refs}

\end{document}